# AI versus AI in Financial Crimes & Detection:
# GenAI Crime Waves to Co-Evolutionary AI

Eren Kurshan
Princeton University
ekurshan@princeton.edu

Dhagash Mehta
BlackRock
dhagash.mehta@blackrock.com

Bayan Bruss
Capital One
bayan.bruss@capitalone.com

Tucker Balch
Emory University
trbalch@gmail.com

**Abstract**: Adoption of AI by criminal entities across traditional and emerging financial crime paradigms has been a disturbing recent trend. Particularly concerning is the proliferation of generative AI, which has empowered criminal activities ranging from sophisticated phishing schemes to the creation of hard-to-detect deep fakes, and to advanced spoofing attacks to biometric authentication systems. The exploitation of AI by criminal purposes continues to escalate, presenting an unprecedented challenge. AI adoption causes an increasingly complex landscape of fraud typologies intertwined with cybersecurity vulnerabilities.

Overall, GenAI has a transformative effect on financial crimes and fraud. According to some estimates, GenAI will quadruple the fraud losses by 2027 with a staggering annual growth rate of over 30% [27]. As crime patterns become more intricate, personalized, and elusive, deploying effective defensive AI strategies becomes indispensable. However, several challenges hinder the necessary progress of AI-based fincrime detection systems. This paper examines the latest trends in AI/ML-driven financial crimes and detection systems. It underscores the urgent need for developing agile AI defenses that can effectively counteract the rapidly emerging threats. It also aims to highlight the need for cooperation across the financial services industry to tackle the GenAI induced crime waves.

## 1. Introduction

The digital transformation of finance and banking has accelerated markedly over the past decade. It was further strengthened by the global COVID-19 pandemic, which solidified digital banking as the foremost channel. This shift has not only revolutionized customer interactions but also profoundly impacted the financial crime landscape. The advent of digital banking has afforded financial criminals unprecedented opportunities to exploit the efficiencies, speed, and vast transaction volumes of digital banking and finance.

In 2023 alone, approximately $3.1 trillion of illicit funds flowed through the international financial infrastructure [43]. The majority of these funds originated from money laundering operations, which marks the final step for a broad range of criminal activities, including an estimated $346.7 billion in human trafficking, $782.9 billion in drug trafficking, and $11.5 Billion from terrorist financing. Also in 2023, global projections indicated that fraud scams and bank fraud schemes led to an aggregate loss of $485.6 Billion [43]. Fraud and financial crimes affect all segments of the population with the worst effects on the most vulnerable segments, women, children and the elderly. Financial crimes also have a serious negative impact on economic development of underdeveloped countries [65].

Following the wide scale adoption and successes of machine learning-based crime detection systems in the past decade, crime groups swiftly adapted AI, unveiling novel crime patterns aimed at targeting a broad range of victims, probing and exploiting the vulnerabilities of detection models. These include a range of tactics from sophisticated GenAI financial scams, to exploiting mobile device vulnerabilities to compromise banking applications, enhanced money laundering operations through more dispersed and harder-to-detect mule networks, exploiting real-time payment systems and to using deep fakes to infiltrate call centers and drain customer accounts.

According to the United Nations Office of Drugs and Crime "organized crime has globalized and turned into one of the world's foremost economic and armed powers" [35]. In 2024, INTERPOL's general secretary expressed "facing an epidemic in the growth of financial fraud, leading to individuals, often vulnerable people, and companies being defrauded on a massive and global scale" [64]. Hence, the rapid adoption of AI by crime organizations has very serious consequences on the future of financial crimes and society.

Meanwhile, AI-based financial crime detection systems face a long list of challenges and obstacles. In 2024, the U.S. Treasury highlighted that "the existing risk management frameworks in financial services may not be adequate to cover emerging AI technologies" [14]. Yet, the latest cost-cutting trends have further impeded the progress in financial crime and fraud detection technologies.

Given the sophistication of the latest AI-driven crimes and the relentless pace of technological advancement, an AI-versus-AI defense appears to be the only viable solution approach moving forward. However, given the numerous organizational and procedural roadblocks and underinvestment in AI-defense systems, it is hard to achieve successful results against the emerging AI-driven financial crime wave.

This paper examines the current state of AI/ML-driven financial crime, focusing on emerging trends and vulnerabilities. By analyzing these factors, it seeks to establish a robust framework for comprehending and addressing the risks associated with AI-enabled financial crimes within today's banking landscape. The paper highlights emerging techniques and strategies to enhance the effectiveness and resilience of defense systems against increasing digital financial threats, aiming to stimulate industry-wide discussions on potential solutions. The paper is organized as follows: Section 2 discusses recent fraud and financial crime trends, Section 3 highlights the limitations of the current financial crime detection system limitations, Section 4 outlines the AI and GenAI opportunities in financial crime detection and prevention, finally Section 5 presents conclusions.

## 2. Recent Fraud & Financial Crime Trends

### 2.1. Crime Volumes, Typologies & Organizations

Since the wide-scale adoption of digital banking, cybersecurity and fraud have become increasingly intertwined. The rise of cyber challenges have resulted in worsening fraud trends in almost all criteria in the past decade [10].

**Volume Increases:** Fraud volumes consistently rose during the global COVID pandemic and concurrent adoption of digital banking [2]. According to [1], fraud value and volumes have climbed by over 50% in most financial institutions. TransUnion reported an 80% rise in digital fraud attempts compared to pre-COVID [3]. Sharp increases were reported in credit card fraud compared to pre-pandemic levels (by 76%), account takeover fraud (ATO) (between 81% [3] to 131% according to [4]), identity theft (by 81%), automated clearing house transactions (ACH) (122%), synthetic ID fraud (132%) and others. Similarly, other studies reported 42% growth in e-commerce fraud in the U.S. during the same period [4].

**Faster Crimes:** Compared to the traditional and slower payment channels, digital banking has driven a strong shift towards real-time and faster payments fraud in recent years. Real-time payments fraud has reached 40% of the fraud volumes in the UK [53]. In 2023, an average of $100 million was sent over the Zelle network every hour, with a year-to-year growth rate of 28% [74]. Though the real-time payment (RTP) adoption is in early stage in the US, fraud growth holds similar risks. The combination of increased volumes and speed pose serious challenges for large payment processors and banks.

**Growing Fraud Losses:** Fraud losses have also been persistently increasing. According to [32], there has been 14% year-to-year growth in the past few years. The FTC reported that fraud losses exceeded $10 Billion in 2023 with a $1 Billion growth from 2022 [37]. Globally, fraud losses exceeded $485 Billion according to 2024 NASDAQ Global Financial Crimes Report [43].

**Diversifying & Increasing Number of Crime Typologies:** Fraud typologies have diversified in almost all domains from account takeover to card-not-present fraud to authorized push payments scams and ID theft. Among the fraud typologies, scams, card-not-present (CNP), cyber fraud, ID-theft have exacerbated and rose to the top consistently over the globe across all geographies [1].

**Rapid Deployment of Novel Crime Typologies:** Novel crime topologies are continuously invented and rapidly implemented. In the past few years investment scams have diversified significantly, as an example, "Rug Pull Scams" emerged in the crypto space involving sudden shuttering of cryptocurrency projects by the developers and causing the investors to lose money.

**Explosion of Financial Scams:** According to the FTC, in 2023 alone, victims lost $2.7 Billion to imposter scams, $1.4 Billion to social media scams, payment scams and cryptocurrency losses were $1.4 Billion. Scams also impacted more young adults (aged between 20-29) than older adults (aged 70+) [39]. Interpol 2024 report highlighted that payment fraud, investment fraud, romance scams and business email compromises are the most widespread financial crime types globally [41]. Government and business impersonation scams also grew in recent years causing the Federal Trade Commission (FTC) to take targeted action [37].

**Rise of Crime-as-a-Service & Scam Call Centers:** The rise and globalization of global crime organizations have caused more sophisticated business models and organizational constructs. One of the most worrisome observations from the Interpol report is the rising prominence of *Crime-As-A-Service* (CAAS) and organizational constructs like *Scam Centers*. This globally distributed network of crime centers are being powered by technology/AI and victims trapped by human trafficking groups [40][41].

The Financial Action Task Force (FATF) highlighted that *Money-Laundering-As-A-Service* also emerged as a new business model with a range of service quality offered to clients based on their pay-grade [43]. For instance, during the layering stage, less frequently used accounts and smaller amounts distributed across a broader network of money mules are provided to clients with high-paying service grades. Rapid adoption of emerging technologies and AI appear to be the unifying force across the different business models and organizational constructs for crime as a business. Similarly, since the introduction of LLMs

*Jailbreaking-As-A-Service* business models emerged rapidly for GenAI-driven financial crimes.

## 2.2. Harder-to-Detect, Dispersed & Faster Crime

Fraud in the digital payment channels is becoming increasingly multi-channel, subtle and harder to detect. As financial institutions struggle with massive amounts of data, criminals use the large data volumes and data types to their advantage to devise harder to detect fraud schemes. While cross-channel feature extraction, integration of multi-channel data still pose challenges to financial institutions and payment processors, criminal groups leverage AI in almost every stage of financial crime. This starts with the reconnaissance stages of thoroughly profiling AI-based crime detection and cyberdefense systems, to optimizing the attack patterns to evade detection. Furthermore, massive data breaches over the past decade on hundreds of millions of customers provided criminal groups the opportunity to combine the data with recent AI models towards developing more customized attacks and to exploit vulnerabilities.

## 2.3. GenAI-based Crime Typologies

In addition to improving the efficiency of existing fraud and financial crime types, the rapid progress in generative AI has also introduced novel crime and scam topologies in the past years. Transformer-based large language models, text to image and video generators have been utilized by criminals to generate language, video, images and other data for a range of crimes from synthetic identity fraud to advanced phishing and smishing attacks, and biometric spoofing attacks targeted towards call center biometrics [12] [5]. Deepfakes have reportedly been surging by around 7 fold in fintech according to some reports [28].

**Voice Cloning:** The advances in GenAI enabled cloning voices from a few seconds of recordings. This progress has fundamentally challenged the voice biometrics authentication landscape [7]. The advances are so risky that some providers decided to not release the voice cloning tools, calling them "too effective to be released to the public" [8]. These capabilities caused a large number of scams to be devised using voice/face/video cloning [30]. However, despite the concerning scam and fraud trends the adoption of voice biometrics in financial services has not slowed down [13]. Voice biometrics solutions are still being used as a part of multi-factor authentication in a wide range of financial service provider systems.

**Social Engineering via Phishing, Smishing, Vishing:** AI-based social engineering tactics through email (phishing), SMS and text messages (smishing), and phone, VoIP, robocall (vishing) have flooded the market in recent years [44]. While typos and grammar errors were considered the most significant giveaways in phishing attacks in the past, the use of generative AI has resulted in perfectly drafted, well-written phishing emails over the past few years. Significant increases have been observed in all social engineering attacks with vishing attacks leading the pack with reportedly 5x growth in 2022 and later in 2023 [45], with 1 out of 4 adults being impacted and 75% of those exposed to such attacks being victimized [46]. Business email compromises (BEC) have also been growing rapidly. A staggering 15+ fold growth was reported [29]. These alarming numbers highlight the need for change in the GenAI powered financial crime era.

**Video Cloning:** Reports of high-quality deep fakes being offered around $150 on the dark web is alarming for the growing risk of video cloning scams [48]. Though not as common yet, the use of video cloning is expected to impact financial services by spoofing virtual meetings with financial advisors and through the ever-growing list of scams. AI-based voice cloning detection techniques were developed in response, yet their effectiveness has severely been challenged with each new generation of GenAI [31].

**Synthetic Identity Fraud:** GenAI-produced contents (such as documents, profiles, text and photos) have been used in synthetic identity fraud. The generated documents can be utilized for a large number of crime types ranging from money laundering to the deep fake documents and synthesized photos in deed fraud [6]. The volumes, sophistication and losses tied to synthetic ID fraud are expected to grow with each generation of AI [67]. Traditional countermeasures like the use of biometrics, third party data are challenged by rapidly advancing AI capabilities.

**Wide Range of AI-Based Scams:** According to Barclays, over 70% of the scams originate on social media, dating applications and digital marketplaces. IC3 report highlights that elderly scams caused close to $3.5 Billion losses in 2023 with a year-to-year increase of around 11%. GenAI has made the elderly scams such as parent/grandparent, attorney, romance and others broadly applicable to the younger victims [17]. A large number of kidnapping scams have been reported in the past year alone with AI-generated images [47]. Video calls from deep-faked family members, officials, financial firm employees, and financial advisors have become harder to detect. Armed with breached data acquired from the dark web, scammers appear more believable for a wide range of emerging scam types [42].

## 2.4. Malicious LLMs & Weaponized AI

Weaponization of GenAI capabilities has become an increasing concern [18]. In June 2023, reports emerged on *WormGPT*, as a dark version of *ChatGPT* capable of a wide range of illicit activities. Similarly, *FraudGPT* was reported as an LLM for finding the system vulnerabilities, writing malware and other malicious code, writing effective phishing emails and other capabilities [19]. Other jailbroken LLMs like *BlackHatGPT, EscapeGPT,*

*LoopGPT* as well as Jailbreaking-As-A-Service also were rapidly introduced since the introduction of ChatGPT-4.

## 2.5 Growing Cybersecurity Vulnerabilities

Globally, cybersecurity and data breaches take the first place in the primary challenge list for fraud and financial crime detection [1]. While social engineering capabilities of GenAI and specifically large language models have garnered a lot of attention [20], AI capabilities in identifying computing system vulnerabilities and exploiting such vulnerabilities have not been studied as rigorously.

**Single-Point Failure through Mobile Device Security:** Today, financial institutions depend heavily on mobile devices and mobile banking applications for digital banking. Mobile banking not only provides ease of access and speed, it reduces the operational costs for financial firms. The growing use of mobile banking resulted in a large number of ATM and branch closures.

Mobile device reliance spans the full range of processes from end-point user authentication to the common use of one-to-passwords for mobile banking authentication. Yet, mobile device security has been more questionable than ever. This centralized power causes single point failure risks for financial institutions and customers, and favorable return-on-investment for criminal groups to invest heavily into viruses, malware, and zero-click attacks on mobile devices [73].

**Zero Day Attacks:** Lately, a number of critical zero-day attacks have been reported impacting millions of customers around the world [21]. Not only were zero day attacks shown to be more common than initially thought, over 60% of the vulnerabilities identified were not known zero-day vulnerabilities [21]. Reports on emerging spyware showed that mobile phones are nowhere close to the security levels required to perform the high-stakes financial services they are used for [22], [23].

**One-Time-Passwords (OTP):** Financial institutions rely heavily on one-time-passcode based authentication (OTP). As mobile device and OTP usage became more prominent, SIM hijacking vulnerabilities also became more wide-spread [24]. As SMS message security primarily depends on cellular network security, the underlying vulnerabilities of the network infrastructure come into question. Lately, these processes have been heavily attacked and are no longer considered secure [25]. Further, specialized mobile phone trojans were developed and reported in SMS-based OTPs. Further, newly released OTP bots have been shown to be initiated through phishing, malware, SMS attacks, and use initial information that may have been acquired through mass compromises and other channels.

**Account Takeover Fraud (ATO):** Both cybersecurity vulnerabilities and social engineering tactics are frequently geared towards account takeover fraud, which has been growing rapidly in the past few years [66]. Due to its versatility and returns (by draining the victims accounts fully), crime groups have strongly invested in ATO in recent years. ATO is also strongly tied to identity theft, money laundering and other crime types. Though behavioral profiling through AI/ML models, third party data and biometrics have helped detect anomalous patterns, criminals have been developing countermeasures like extensive account priming tactics, targeted malware and other techniques to circumvent the AI defenses.

## 2.6 System-Level AI Vulnerabilities

The 'Market Bombing' attack of May 2023 highlighted the persistent vulnerabilities of the financial AI systems to AI-based attacks. During this attack, an AI-generated image of an explosion started circulating in social media. The original post was not only boosted by bots but was reportedly retweeted by verified news channels adversarially. AI-based social media tracking systems picked up the signal and started shorting the stocks, which resulted in a downward spiral. The Dow Jones Industrial Average fell about 80 points over a 4 minute period [15].

This attack type highlights the vulnerabilities of large scale AI systems to AI-based attacks. In building complex applications, many systems use a multitude of subsystems and AI-models with complex interactions. As a result, it is highly difficult to fully profile under all possible circumstances with the current technology and hence ensure safety of such systems. Furthermore, the AI building blocks carry a considerable amount of unpredictability and risk that is hard to safely manage at this point.

## 2.7 AIs Unlimited Crime Potential

AIs unlimited crime potential was highlighted by a number of studies, with potential crimes ranging from AI manipulation based attacks, to AI-based attacks on infrastructure and weaponization of AI systems [9]. Increased autonomy and growing complexity worsens the outlook. With the right reinforcement and incentive structure AI may not need humans to invent new crime tactics, and even new crime types. Though considered futuristic by some, AI systems have demonstrated the capability to invent and adopt unethical behaviors based on incentives even in financial settings. Current infrastructure defense solutions are nowhere close to being capable of dealing with the upcoming AI-driven crime waves.

## 3. Financial Crime Detection Limitations

### 3.1 Criminal Organizations Adopt AI Faster

Recent reports show that AI is heavily adopted and deployed by criminals at a much faster rate than financial service providers. The gap in technology adoption is a fundamental challenge in the future of financial crime detection as AI is becoming increasingly more capable of sophisticated attacks and financial crimes. On the flip side, the decreased investments in the financial crime detection systems creates an uneven playing field favorable for criminals.

## 3.2 Model Agility & Model Governance

While criminal organizations rapidly deploy novel crime typologies and scams, financial services firm AI teams often deal with long model governance processes. MRM processes often involve not only model development and documentation periods, but numerous committee reviews like stakeholder reviews/approvals, compliance and risk reviews in addition to the core model risk management reviews and processes [16]. These steps frequently take months and in some cases even years. They unintentionally hurt model agility and create significant vulnerabilities. On the other hand, rapid adoption of AI/ML models also carries strong risks and compliance issues, which is fundamentally a more delicate optimization problem for financial firms. Modernization of MRM processes for AI and GenAI is an essential next step to stop the developing AI-induced crime wave.

## 3.3 Organizational & Cultural Challenges

In general, the adaptation of AI for financial crime detection is significantly higher in retail banking as the fraud losses often directly translate to the firm's bottom-line. For investment management firms, unlike retail banking, financial crime programs fall into the compliance scope and the losses have an indirect impact on the firms. As a result, even outdated rule-based systems continue being deployed in such segments, with limited investment in AI/ML-based systems and hiring of technical experts.

## 3.4 Limited AI/ML & Talent Investment

Crime groups strongly invested in AI and technology in the past decade, which caused their emerging technological organizations to become *Criminal Silicon Valleys* [49]. In contrast, recent cost cutting trends have strongly impacted AI/ML investments in the financial crime detection space.

**Limited AI/ML/Technology Hiring:** The lack of investments in training and acquisition of new talent in financial institutions has even been sharper [27]. The perception of technology (even AI/ML) as a *low-end commodity* in some financial firm cultures often results in untrained back-office personnel leading technology strategy and implementation roles instead of trained technologists, causing ineffective financial crime detection and defense systems with heavy reliance on rules, few and ineffective features and high false positives. Similarly, in such technology-averse cultures technology strategy development and technical decisions are led by untrained back-office business personnel yielding unsuccessful outcomes.

**Low-End Investment in AI:** Financial services industry frequently exhibits underinvestment in emerging technologies for financial crime detection. This underinvestment ranges from deficiencies in hiring skilled technologists to understaffing and undercompensating crime detection teams. This trend, in turn, has created and continues to create an ideal scenario for criminals to grow their illicit footprint. While some banks are more technology embracing and publicly state their high-investment levels in technology to prevent cyber crimes and fraud, others have limited to no investment [48].

**Back-Office Labeling of Crime Detection:** Notably, even today, the reliance on purely rule-based solutions persists in some segments, driven by a reluctance to commit resources for AI and technology investments comparable with those allocated to traditional front-office functions. Moreover, within the banking sector, AI and technological solutions remain predominantly relegated to back-office operations, with significantly lower investment compared to front-office initiatives.

## 3.5 Lack of Industry-Wide Cooperation

The siloed nature of individual authentication systems in financial firms enables criminals to attack multiple banks with the same attack vectors. As financial firms are not incentivized to go public with such attacks, crime groups can perpetrate the same attack types repeatedly. In order to prevent this, firms were encouraged to share information and collaborate against the rising AI risks by the government and regulators [14]. Forming collaboration and cooperation against growing criminal organization capabilities are essential in the AI-driven crime era. Collaboration and cooperation opportunities may range from federated learning of crime patterns across different financial institutions [58], to data sharing about suspicious entities and emerging crime patterns.

## 3.6 GenAI Cost & Risk Challenges

Even though GenAI and LLMs provide numerous opportunities in financial crime detection, their development and run-time costs have been rising in each generation. The exponential growth in the number of parameters has caused models with trillion parameters to hit the market lately [51]. As a result, development costs exceeding $100 Million and daily run-time costs around $1 Million have been reported [52]. Though the scaling trends have been backtracked by some technology firms, the AI development, deployment and risk-management costs discouraged a large number of financial institutions from pursuing generative AI-based solutions.

**GenAI Risks:** Despite their growing use in numerous application domains, LLMs carry various risks and compliance challenges that range from hallucinations [55], data leakage [54], jailbreaking [56], security and privacy vulnerabilities [57] like poisoning attacks [61], backdoors [62] that can have catastrophic consequences. As an example, the use of retrieval-augmented generation models (RAG) in financial use cases to achieve up-to-date, accurate outputs with source attribution can be particularly risky due to poisoning attacks [63]. For regulated industries with heavy emphasis on accuracy such as financial services, it is harder to adopt compliant genAI solutions with manageable risk ranges, accuracy guarantees, devoid of hallucinations and falsehoods.

## 4. Opportunities

### 4.1 GenAI for Crime Detection

Despite enabling novel crime typologies, generative AI also provides a wide range of opportunities in financial crime and fraud detection. GenAI-based solutions are being built and deployed for advanced transaction monitoring, behavioral profiling, in building effective and faster know-your-client (KYC) solutions, negative news (NN) tracking for politically-exposed-persons (PEPs), internal crime and fraud detection through communications monitoring, end-to-end AML processes and others [36].

### 4.2 GraphAI/ML for Fraud & Crime Detection

Despite the criminal AIs' improved capabilities in concealing crimes from traditional machine learning models, graph AI/ML provided key solution avenues in financial crime detection. Instead of focusing on the individual accounts, transactions, and behavioral patterns that can be increasingly mimicked by AI, these solutions focus on a network of entities (individuals, accounts, transactions, devices etc) and capture higher-level patterns beyond the individual entities. This yields improved crime detection rates for a wide range of crimes from mobile banking to money laundering steps.

GraphAI/ML solutions identify anomalous patterns through network analysis, flow analysis that follow money movement over the entire network instead of focusing on individual transactions, propagate risk across a range of entities through connectivity over the graphs, predict previously unknown connections through graph completion and inference. They are capable of learning at various levels of granularity over large networks [34]. Further, integrating deep learning solutions with retrieval augmented generation capabilities (RAGs) and knowledge graphs can enable further capabilities in continuously learning complex, temporal, and probabilistic relationships for crime and anomaly detection [50].

### 4.3 Multi-Channel, Multi-Modal AI/ML for Fraud

Multi-modal and multi-channel AI models can also be leveraged by training on fraud and financial crime data (e.g. natural language to transaction data) to detect specific crime instances dispersed across multiple channels.

### 4.4 Adoptive & Dynamic AI

Dynamically adapting to the changing crime and data patterns as well as emerging risks is an important feature in AI-based advanced financial crime detection solutions. As an example, the emergence of deep fakes can be mitigated by adjusting and eliminating the weight of the corresponding features in multi-modal authentication as well as modulating the signals out of the corresponding authentication systems as risk and crime patterns emerge [26]. Similarly, data risk can be factored into authentication solutions continuously to adjust for the emerging trends.

### 4.5 AI-Based Cyber Defense for AI

In recent years large language models and broadly NLP systems have experienced a wide range of emerging attack types [72]. Given the proliferation of targeted cyberattacks aimed at AI systems, developing effective defense systems for the financial services AI is an absolute necessity [70]. As crime groups continue investing in AI, custom techniques to protect the AI-based crime detection solutions from evasion, profiling, poisoning, privacy, abuse and other adversarial attacks is of prime importance [71].

### 4.6 Data Sharing & Federated Learning

As highlighted by the Treasury, industry-wide defenses, collaboration and data sharing to prevent financial crimes and fraud can be important defense assets [14]. Governments and regulators may play an important leadership and facilitator role in such efforts. As an example, 6 banks and the Monetary Authority of Singapore announced data sharing agreements towards money laundering and terrorist financing in 2024, after the introduction of the 2023 Bill [59]. Privacy preserving federated learning of crime patterns across multiple financial firms also has the potential to help reduce emerging fraud risks [58]. Small scale federated projects started for anti-money laundering applications in recent years as in [60].

### 4.7 Risk-Aware AI Systems & Authentication

Given the effectiveness of AI in generating publicly available data such as (image, video, voice, speech), alternative biometric authentication solutions with more augmented visual perception have been proposed. These techniques include harder to reach biometrics to enhanced spectrum biometric authentication and multi-factor solutions [33]. Similarly, eliminating or risk-based modulating high-risk biometric features easily accessible to criminals through risk-aware authentication and cybersecurity frameworks are promising approaches [26].

## 4.8 AI Regulations & Self-Regulatory AI

Recent regulatory updates and standards have shown that regulations play an important role in preventing fraud. In Europe, stricter payment fraud security and customer authentication requirements introduced by the Payment Services Directive (PSD2) for card-not-present transactions, along with the implementation of industry standards for chip cards to combat card-present fraud, have substantially decreased instances of card fraud [11]. This underscores the crucial role of regulations in addressing fraud. Newly introduced AI regulations in the U.S. [68] and Europe [69] try to fill the current gaps and require financial institutions to adapt to the rapidly changing AI regulation landscape.

Self-regulating and regulatory AI appear to be essential in eliminating long and human-based model governance processes and monitoring reviews to keep up with rapidly evolving criminal AI systems [16]. Using both regulatory AI systems and self-regulation capabilities in models, as well as real-time and continuous monitoring through regulatory AI enables the required agility in AI-based cyber and fraud-defense systems while providing mechanisms for regulatory compliance and risk management.

## 4.9 Co-Evolutionary AI

Enabling financial crime detection systems to co-evolve with AI-based crime models may be a viable solution opportunity to appropriately address the growing threats of AI-driven financial crimes. While co-evolutionary approaches have been considered for problem solving in multi-agent systems, they naturally lend themselves to financial crime AI and AI-based detection systems.

## 5. Conclusions

Generative AI has shown its transformative power in financial crimes in the past few years. As crime groups continue embracing emerging technology and AI capabilities, the financial services industry needs to take swift action in developing appropriate responses to the growing challenges. In order to address the alarming AI-crime trends, financial firms are expected to invest more into AI-based crime detection solutions and embrace forward-looking AI strategies. Modernizing MRM processes for GenAI is critical in enabling more agile AI systems for crime detection. This can enable co-evolutionary scenarios, where AI-based financial crimes can be properly addressed by rapidly evolving AI-based financial crime detection systems.